# Human behaviour through a LENS

**How Linguistic content triggers Emotions and Norms and determines Strategy choices**

Valerio Capraro

University of Milan-Bicocca

valerio.capraro@unimib.it

**Abstract**

Over the last two decades, a growing body of experimental research has provided evidence that linguistic frames influence human behaviour in economic games, beyond the economic consequences of the available actions. This article proposes a novel framework that transcends the traditional confines of outcome-based preference models. According to the LENS model, the Linguistic description of the decision problem triggers Emotional responses and suggests potential Norms of behaviour, which then interact to shape an individual's Strategic choice. The article reviews experimental evidence that supports each path of the LENS model. Furthermore, it identifies and discusses several critical research questions that arise from this model, pointing towards avenues for future inquiry.

# Introduction

Understanding human behaviour in economic games has been a major area of study for decades. In particular, one-shot and anonymous interactions have garnered significant attention, because they allow to study human behaviour in its purest form, undistorted by potential future consequences or social influences.

It has long been known that in such contexts, individuals do not merely maximize their monetary outcome. For instance, a substantial proportion of people share their money in the dictator game (Kahneman, Knetsh & Thaler, 1986). This raises the question: if not material gain, what utility are individuals seeking to maximize?

The search of this utility function has been a central focus of research. An influential line of work pertains to "social preferences". While differing in many details, these models are based on a foundational assumption: that a player's utility depends only on the monetary payoffs of all individuals involved in the interaction. For instance, Ledyard (1995) postulates that a decision-maker's utility depends not just on their monetary payoff, but also on the sum of the monetary payoffs of the other people involved in the interaction. Fehr and Schmidt (1999) assume that individuals care about reducing economic differences. Charness and Rabin (2002) posit an inclination to increase total welfare. See Capraro and Perc (2021) for a review.

Yet, this "consequentialist assumption" – the foundational assumption that decisions are purely based on monetary consequences – is facing increasing criticism.

## The effect of linguistic content

A major criticism emerges from experiments emphasizing the impact of linguistic content on individuals' choices.

Liberman et al. (2004) found that when the name of a Prisoner's dilemma was altered from the "Wall Street game" to the "Community game", individuals became more inclined to cooperate. Eriksson et al. (2017) presented participants with an ultimatum game. The action of declining a proposer's offer was labelled differently in two scenarios: as "rejecting the proposer's offer" and "reducing the proposer's payoff". Despite the monetary equivalence of these two scenarios, responders demonstrated a higher tendency to decline low offers when confronted with the term "rejection". Capraro and Rand (2018) observed participants' choices between equitable and efficient money distributions. By merely tweaking the description of each choice – labelling one as the "nice thing to do" – they revealed that individuals typically chose the positively framed option, irrespective of its actual implications. Capraro and Vanzo (2019) conducted six dictator game treatments, with differing instructions. For example, in the "boost" treatment, the altruistic action was labelled as "boosting" the recipient. Once again, linguistic frames led to significant behavioural variations. Subsequent research has corroborated these findings (Chang et al., 2019; Huang et al., 2019; Huang et al., 2020; Mieth et al., 2021; Kuang & Bicchieri, 2024).

Some studies have also unveiled a potential dark side to this phenomenon. Capraro et al. (2022) revealed that when dictator game receivers could select the game's linguistic frame, they opted for the terminology likelier to yield higher personal payoffs. Ścigała et al. (2022) found that individuals with high Honesty-Humility traits could be enticed into accepting a bribe if it was portrayed as a "cooperation act". In essence, linguistic frames can be manipulated by self-interested parties to their advantage.

These findings challenge the consequentialist assumption that utility functions solely depend on monetary outcomes. Instead, they emphasize the critical role of linguistic frames, and the need of a "paradigm shift from outcome-based to language-based preferences" (Capraro et al., 2024b). A central inquiry within this shift is: How do linguistic frames affect people's decisions?

**The LENS model**

The LENS model introduces a new framework for understanding human behaviour, moving beyond the confines of an outcome-based perspective. It aims to describe how linguistic content shape people's decisions.

At its core, the model posits that Language works primarily by evoking certain Emotions and suggesting specific Norms of behaviour within the context at hand. These emotions and norms then interact and determine Strategy choice. See Figure 1 for a qualitative description. Currently, the LENS model is qualitative in nature. A major line of future research consists in its transformation into a quantitative model, by means of a suitable utility function.

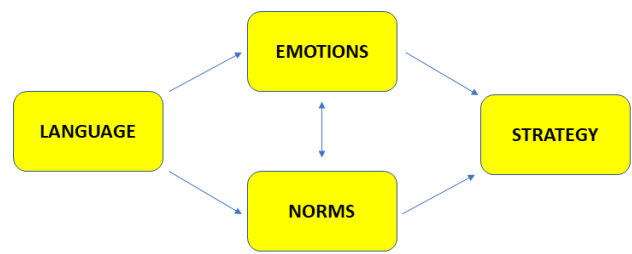

*Figure 1. Qualitative representation of the LENS model.*

The following sections review the evidence for each path of the model and outline key questions for future work.

**The language - emotions path**

The impact of language on emotions is evident in everyday life. Descriptive prose in literature can evoke a spectrum of emotions ranging from joy and sorrow to love and anger. However, while this qualitative relationship is widely acknowledged, a challenging question arises: How can we quantify the emotions triggered by a specific piece of text?

Sentiment analysis offers a promising approach. At its heart, sentiment analysis tools assess textual content to determine its sentiment tenor. Earlier tools predominantly categorized emotions as "positive" or "negative" (e.g., Sebastiani & Esuli, 2006). More recent tools have surpassed this binarism and associate words with different emotions (e.g., Mohammad & Turney, 2013). While sentiment analysis has found applications across various sectors (e.g., Bollen, Mao, & Zeng, 2011; De Choudhury, Gamon, Counts, & Horwitz, 2013; Tumasjan, Sprenger, Sandner, & Welpe, 2010), there have been limited attempts to harness it to explain people's behaviour in economic games.

This application is not straightforward, as illustrated by a pilot study by Capraro and Vanzo. In the dictator game, the study found that participants' self-reported sentiments changed based on the terminology used. However, one of the renowned sentiment analysis tools, SentiWordNet, failed to account for these variations, even at a qualitative level.

The primary limitation of this tool lies in its limited context dependency. While SentiWordNet does offer a degree of context-dependency — a word can belong to different "synsets" (sets of synonyms) and, thus, can convey different sentiments based on context — it was not nuanced enough to capture the observed differences in the study. See Table 1.

| Word | SentiWordNet score | Actual score |
|---|---|---|
| Boost | 0.25 | 0.25 |
| Give | 0 | 0.325 |
| Donate | 0.625 | 0.355 |
| Demand | -0.25 | -0.34 |
| Take | 0 | -0.315 |
| Steal | -0.5 | -0.49 |

*Table 1*. SentiWordNet scores were computed using the synsets: steal#1 = "take without the owner's consent"; take#8 = "take into one's possession"; demand#1 = "request urgently and forcefully"; give#3 = "transfer possession of something concrete or abstract to somebody"; donate#1 = "give to a charity or good cause"; and boost#2 = "be beneficial to". Actual scores were asked to a sample of 567 subjects living in the USA, recruited through Amazon Mechanical Turk.

Consequently, a crucial inquiry is: How can sentiment analysis be refined to capture context more effectively, especially to better understand behaviour in economic games?

Recent research has begun addressing this question. Building on work by Rathje et al. (2023), who reported that GPT if an effective tool to conduct sentiment analysis, Capraro et al. (2024a) showed that context-dependent sentiment scores estimated by GPT-4 explain human behaviour in the dictator game, beyond the economic consequences of the available actions.

**The language - norms path**

Several studies have highlighted the ability of language to influence people's perceptions of norms within specific contexts. Eriksson et al. (2017) revealed that participants deemed "reducing the proposer's payoff" as more morally acceptable than "declining the proposer's offer". Capraro and Rand (2018) demonstrated that individuals' moral judgments in dilemmas pitting equity against efficiency depended on the linguistic frame of the game. Capraro and Vanzo (2019) found that the label assigned to the available actions significantly affected moral judgments in dictator games. This body of research shows that linguistic content significantly affects personal norms, internal beliefs about right and wrong. Linguistic frames affect also injunctive norms, people's perception of what others believe to be socially appropriate (Chang et al., 2019). It is plausible that suitable linguistic content may influence also people's beliefs about others' behaviour (descriptive norm), although there is little research on this specific topic.

A major question in this area pertains to the quantification of the normative value of text. Could the methods of sentiment analysis be adapted to develop a "normative analysis"? This question is fascinating. A normative analysis should ideally associate each text with multiple scores reflecting "fundamental moral values". Therefore, this question is linked to a foundational question in moral psychology: What constitutes these fundamental moral values? This question is still debated, with at least two theories, moral foundations theory (Graham et al., 2009) and morality-as-cooperation theory (Curry et al., 2019). There have been some attempts in creating text corpuses that might be the starting point for computational models to detect moral values in text (Hoover et al., 2020; Trager et al., 2022; Preniqi et al., 2024).

**The emotions - strategy path**

Numerous studies have underscored the link between emotions and human behaviour. Earlier studies have shown a causal link between specific emotions and decisions. Strohminger et al. (2011) and Ugazio et al. (2012) highlighted that anger often amplifies consequentialist judgments in moral dilemmas; in contrast, disgust increases deontological judgments. Motro et al. (2018) revealed that anger and guilt may have contrasting effects in dishonesty tasks. Participants tasked with recalling angry experiences exhibited an increased propensity to lie, while those remembering guilt-ridden memories demonstrated heightened honesty.

Some studies have adopted a broader dual-process perspective and explored whether generally inducing people to "rely on emotions" has an impact on decisions. It has been found that relying on emotions increases cooperation in the prisoner's dilemma (Levine et al., 2018), reduces instrumental harm for the greater good (Capraro et al., 2019) and the extent to which people believe humans are superior to other animals (Caviola & Capraro, 2020).

Although these results provide compelling evidence that emotions may affect behaviour, a myriad of important questions remain unanswered. In particular, a classification of links between emotions and behaviours is missing. Which behaviours predominantly stem from emotions as opposed to deliberation? And, amongst the emotionally-driven behaviour, which specific emotions drive which behaviours?

**The norms - strategy path**

The idea that individual behaviour is influenced by perceived norms is well-accepted (Cialdini et al., 1990; Bicchieri, 2005; Bicchieri et al., 2022; Dimant & Shalvi, 2022; Hertwig & Mazar, 2022; Zickfeld et al., 2023). A substantial body of research has sought to identify the specific norms that propel certain behaviours. When considering one-shot and anonymous interactions, numerous studies suggest that personal norms may play a critical role. Schwartz's (1977) seminal article, for instance, demonstrated a significant association between personal norms and altruistic behaviour, with this association being greater for subjects whose norm was more stable over time. This association between personal norms and behaviour in the dictator game has been confirmed in more recent studies (Capraro & Vanzo, 2019; Capraro et al., 2019). Additionally, Capraro & Rand (2018) illustrated how personal norms can influence the equity-efficiency trade-off.

Other studies provided evidence that injunctive norms may be associated with behaviour in the dictator game (Krupka & Weber, 2013). However, the relative influence of personal, injunctive, and descriptive norms remains unclear. Catola et al. (2021) found that in the one-shot and anonymous public good game, personal norms explain behaviour to a greater extent than social norms. Bašić and Verrina (2021) found that in anonymous dictator, ultimatum, and public good games, personal norms explain behaviour to a greater extent than injunctive norms. Yet, when choices are made publicly, injunctive norms seemed to wield influence comparable to personal norms. It is plausible that social norms assume greater significance in public decisions due to heightened reputational concerns.

An intriguing aspect to explore is the role of descriptive norms. In a trade-off game conducted by Capraro and Rand (2018), participants were presented with two contrasting norms: a personal norm and a descriptive norm. In this context, individuals leaned towards the personal norm. Nonetheless, in situations where decisions may depend on beliefs about others' behaviours (like in the prisoner's dilemma or public good game), descriptive norms are likely to play a more important role.

Some studies have also reported null effects of nudging social and personal norms on honest behaviour (Dimant et al., 2020; Huynh et al., 2024), or even backfire effects, such that nudging personal norms may lead to more self-serving behaviour in a task where participants were tempted to overcompensate themselves after completing an effort task (Morvinski et al., 2023). These studies underscore that the influence of norms might be highly context dependent.

In summary, the central query in this domain remains: How do diverse norms influence behaviour across varied contexts?

**The emotions - norms interaction**

Emotions and norms do not operate in isolation but rather interact. According to moral foundations theory (Graham et al., 2009; Atari et al., 2023), people's moral values are not just rationally determined but are deeply rooted in moral intuitions, which are inherently supported by potent moral emotions. Extending this perspective, one can argue that collective emotional responses might amalgamate to create or reinforce social norms. For instance, widespread social outrage or empathy towards certain issues can set or redefine the acceptable behaviours in a community.

Conversely, established norms can evoke specific emotions in individuals. Consider the experience of a tourist who, unfamiliar with local customs, inadvertently breaches a social norm, such as walking on Amsterdam's bike lanes. Such a faux pas, especially if pointed out publicly, could elicit powerful feelings of shame or guilt. These emotions can act as potent deterrents, ensuring the individual remains compliant with local norms in the future.

Yet, the interplay between emotions and norms remain largely uncharted in academic research. How, then, do emotions and norms interact to shape decisions?

**Conclusion**

A growing body of evidence suggests that linguistic content significantly impacts decisions beyond monetary outcomes. This article introduces a framework to understand human behaviour, moving past the narrow focus of outcome-based preferences. According to the LENS model, Linguistic content evokes Emotions and suggests Norms, which then interact to influence Strategy choice. The article examines evidence supporting each path of the model and raises critical questions for future research. Ultimately, this article aims to contribute to the evolution of behavioural modelling by recognizing and underscoring the critical role of language in shaping human actions.

**Conflict of interest statement**

The author declares no conflict of interest.

# References

*References of particular interest have been highlighted as*
*\* \* of outstanding interest*
*\* of special interest*

*Atari, M., Haidt, J., Graham, J., Koleva, S., Stevens, S. T., & Dehghani, M. (2023). Morality beyond the WEIRD: How the nomological network of morality varies across cultures. *Journal of Personality and Social Psychology*.

**Bašić, Z., & Verrina, E. (2021). Personal norms—and not only social norms—shape economic behavior. *MPI Collective Goods Discussion Paper*, (2020/25).

Bicchieri, C. (2005). *The grammar of society: The nature and dynamics of social norms*. Cambridge University Press.

*Bicchieri, C., Dimant, E., Gächter, S., & Nosenzo, D. (2022). Social proximity and the erosion of norm compliance. *Games and Economic Behavior*, *132*, 59-72.

Bollen, J., Mao, H., & Zeng, X. (2011). Twitter mood predicts the stock market. *Journal of Computational Science*, *2*, 1-8.

**Capraro, V., Di Paolo, R., Perc, M., & Pizziol, V. (2024a). Language-based game theory in the age of artificial intelligence. *Journal of the Royal Society Interface*, *21*, 20230720.

Capraro, V., Everett, J. A., & Earp, B. D. (2019). Priming intuition disfavors instrumental harm but not impartial beneficence. *Journal of Experimental Social Psychology*, *83*, 142-149.

**Capraro, V., Halpern, J. Y., & Perc, M. (2024b). From outcome-based to language-based preferences. *Journal of Economic Literature*.

Capraro, V., Jagfeld, G., Klein, R., Mul, M., & de Pol, I. V. (2019). Increasing altruistic and cooperative behaviour with simple moral nudges. *Scientific Reports*, *9*, 11880.

*Capraro, V., & Perc, M. (2021). Mathematical foundations of moral preferences. *Journal of the Royal Society Interface*, *18*, 20200880.

Capraro, V., & Rand, D. G. (2018). Do the right thing: Experimental evidence that preferences for moral behavior, rather than equity or efficiency per se, drive human prosociality. *Judgment and Decision Making*, *13*, 99-111.

Capraro, V., & Vanzo, A. (2019). The power of moral words: Loaded language generates framing effects in the extreme dictator game. *Judgment and Decision Making*, *14*(3), 309-317.

*Capraro, V., Vanzo, A., & Cabrales, A. (2022). Playing with words: Do people exploit loaded language to affect others' decisions for their own benefit?. *Judgment and Decision making*, *17*(1), 50-69.

**Catola, M., D'Alessandro, S., Guarnieri, P., & Pizziol, V. (2021). Personal norms in the online public good game. *Economics Letters*, *207*, 110024.

Caviola, L., & Capraro, V. (2020). Liking but devaluing animals: Emotional and deliberative paths to speciesism. *Social Psychological and Personality Science*, *11*, 1080-1088.

Cialdini, R. B., Reno, R. R., & Kallgren, C. A. (1990). A focus theory of normative conduct: Recycling the concept of norms to reduce littering in public places. *Journal of Personality and Social Psychology*, *58*, 1015.

Chang, D., Chen, R., & Krupka, E. (2019). Rhetoric matters: A social norms explanation for the anomaly of framing. *Games and Economic Behavior*, *116*, 158-178.

Charness, G., & Rabin, M. (2002). Understanding social preferences with simple tests. *The Quarterly Journal of Economics*, *117*, 817-869.

Curry, O. S., Mullins, D. A., & Whitehouse, H. (2019). Is it good to cooperate? Testing the theory of morality-as-cooperation in 60 societies. *Current Anthropology*, *60*, 47-69.

De Choudhury, M., Gamon, M., Counts, S., & Horvitz, E. (2013). Predicting depression via social media. In *Proceedings of the international AAAI conference on web and social media* (Vol. 7, No. 1, pp. 128-137).

Dimant, E., Van Kleef, G. A., & Shalvi, S. (2020). Requiem for a nudge: Framing effects in nudging honesty. *Journal of Economic Behavior & Organization*, *172*, 247-266.

*Dimant, E., & Shalvi, S. (2022). Meta-nudging honesty: Past, present, and future of the research frontier. *Current Opinion in Psychology*, 101426.

Eriksson, K., Strimling, P., Andersson, P. A., & Lindholm, T. (2017). Costly punishment in the ultimatum game evokes moral concern, in particular when framed as payoff reduction. *Journal of Experimental Social Psychology*, *69*, 59-64.

Fehr, E., & Schmidt, K. M. (1999). A theory of fairness, competition, and cooperation. *The Quarterly Journal of Economics*, *114*, 817-868.

Graham, J., Haidt, J., & Nosek, B. A. (2009). Liberals and conservatives rely on different sets of moral foundations. *Journal of Personality and Social Psychology*, *96*, 1029.

*Hertwig, R., & Mazar, N. (2022). Toward a taxonomy and review of honesty interventions. *Current Opinion in Psychology*, 101410.

Hoover, J., Portillo-Wightman, G., Yeh, L., Havaldar, S., Davani, A. M., Lin, Y., ... & Dehghani, M. (2020). Moral foundations twitter corpus: A collection of 35k tweets annotated for moral sentiment. *Social Psychological and Personality Science*, *11*, 1057-1071.

Huang, L., Lei, W., Xu, F., Yu, L., & Shi, F. (2019). Choosing an equitable or efficient option: A distribution dilemma. *Social Behavior and Personality: An international journal*, *47*, 1-10.

Huang, L., Lei, W., Xu, F., Liu, H., Yu, L., Shi, F., & Wang, L. (2020). Maxims nudge equitable or efficient choices in a Trade-Off Game. *PloS One*, *15*, e0235443.

**Huynh, L. D. T., Stratmann, P., & Rilke, R. M. (2024). No influence of simple moral awareness cues on cheating behaviour in an online experiment. *Journal of Behavioral and Experimental Economics*, *108*, 102126.

Kahneman, D., Knetsch, J. L., & Thaler, R. H. (1986). Fairness and the assumptions of economics. *Journal of Business*, S285-S300.

Krupka, E. L., & Weber, R. A. (2013). Identifying social norms using coordination games: Why does dictator game sharing vary?. *Journal of the European Economic Association*, *11*, 495-524

**Kuang, J., & Bicchieri, C. (2024). Language matters: How normative expressions shape norm perception and affect norm compliance. *Philosophical Transactions of the Royal Society B*, *379*, 20230037.

Ledyard, J, O. (1995). Public goods: A survey of experimental research. In: *Handbook of Experimental Economics*, ed. Kagel J. and Roth A. Princeton, NJ: Princeton University Press.

Levine, E. E., Barasch, A., Rand, D., Berman, J. Z., & Small, D. A. (2018). Signaling emotion and reason in cooperation. *Journal of Experimental Psychology: General*, *147*, 702.

Liberman, V., Samuels, S. M., & Ross, L. (2004). The name of the game: Predictive power of reputations versus situational labels in determining prisoner's dilemma game moves. *Personality and Social Psychology Bulletin*, *30*, 1175-1185.

*Mieth, L., Buchner, A., & Bell, R. (2021). Moral labels increase cooperation and costly punishment in a Prisoner's Dilemma game with punishment option. *Scientific Reports*, *11*, 10221.

Mohammad, S. M., & Turney, P. D. (2013). Crowdsourcing a word–emotion association lexicon. *Computational Intelligence*, *29*, 436-465.

*Morvinski, C., Saccardo, S., & Amir, O. (2023). Mis-nudging morality. *Management Science*, *69*, 464-474.

Motro, D., Ordóñez, L. D., Pittarello, A., & Welsh, D. T. (2018). Investigating the effects of anger and guilt on unethical behavior: A dual-process approach. *Journal of Business Ethics*, *152*, 133-148.

**Preniqi, V., Ghinassi, I., Kalimeri, K., & Saitis, C. (2024). MoralBERT: Detecting Moral Values in Social Discourse. *arXiv preprint arXiv:2403.07678*.

**Rathje, S., Mirea, D. M., Sucholutsky, I., Marjieh, R., Robertson, C., & Van Bavel, J. J. (2023). GPT is an effective tool for multilingual psychological text analysis. *Available at https://osf.io/preprints/psyarxiv/sekf5*

Sebastiani, F., & Esuli, A. (2006). Sentiwordnet: A publicly available lexical resource for opinion mining. In *Proceedings of the 5th international conference on language resources and evaluation* (pp. 417-422). European Language Resources Association (ELRA) Genoa, Italy.

Schwartz, S. H. (1977). Normative influences on altruism. In *Advances in experimental social psychology* (Vol. 10, pp. 221-279). Academic Press.

*Ścigała, K. A., Zettler, I., Pfattheicher, S., & Capraro, V. (2022). Corrupting the prosocial people: does cooperation framing increase bribery engagement among prosocial individuals? Stage 1 Registered Report.

Strohminger, N., Lewis, R. L., & Meyer, D. E. (2011). Divergent effects of different positive emotions on moral judgment. *Cognition*, *119*, 295-300.

Trager, J., Ziabari, A. S., Davani, A. M., Golazizian, P., Karimi-Malekabadi, F., Omrani, A., ... & Dehghani, M. (2022). The moral foundations reddit corpus. *arXiv preprint arXiv:2208.05545*.

Tumasjan, A., Sprenger, T., Sandner, P., & Welpe, I. (2010). Predicting elections with twitter: What 140 characters reveal about political sentiment. In *Proceedings of the international AAAI conference on web and social media* (Vol. 4, No. 1, pp. 178-185).

Ugazio, G., Lamm, C., & Singer, T. (2012). The role of emotions for moral judgments depends on the type of emotion and moral scenario. *Emotion*, *12*, 579.

**Zickfeld, J. H., Ścigała, K. A., Weiss, A., Michael, J., & Mitkidis, P. (2023). Commitment to honesty oaths decreases dishonesty, but commitment to another individual does not affect dishonesty. *Communications Psychology*, *1*(1), 27.